\documentclass[preprint,10pt]{elsarticle}
\usepackage{xcolor}
\usepackage{colortbl}
\usepackage{hyperref}
\usepackage{graphicx}
\usepackage{amssymb}
\usepackage{lineno}
\usepackage{amssymb}
\usepackage{scalerel}
\usepackage{amsmath}
\usepackage{cleveref}
\usepackage{algorithm}
\usepackage[noend]{algpseudocode}
\usepackage[onehalfspacing]{setspace}
\usepackage{wrapfig}
\usepackage{multirow}
\usepackage{tabu}
\usepackage[figurename=Fig.]{caption}
\usepackage{url}

\makeatletter
\def\BState{\State\hskip-\ALG@thistlm}
\makeatother
\newcommand*{\field}[1]{\mathbb{#1}}

\newlength\myindent
\algdef{SE}[SUBALG]{Indent}{EndIndent}{}{\algorithmicend\ }%
\algtext*{Indent}
\algtext*{EndIndent}

\begin{document}
\setstretch{1.0}
\begin{frontmatter}
\title{Measuring the effects of confounders in medical supervised classification problems: the Confounding Index (CI)}

\author[sns,infn]{Elisa Ferrari}
\ead{elisa.ferrari@sns.it}
\address[sns]{Scuola Normale Superiore,
Italy.}

\author[infn]{Alessandra Retico}
\ead{alessandra.retico@pi.infn.it}
\address[infn]{Pisa Division, INFN,
Italy.}

\author[diunipi]{Davide Bacciu}
\ead{bacciu@di.unipi.it}
\address[diunipi]{Dipartimento di Informatica,
Universit\`a di Pisa,
Italy.}

\begin{abstract}
Over the years, there has been growing interest in using Machine Learning techniques for biomedical data processing. When tackling these tasks, one needs to bear in mind that biomedical data depends on a variety of characteristics, such as demographic aspects (age, gender, etc) or the acquisition technology, which might be unrelated with the target of the analysis.
In supervised tasks, failing to match the ground truth targets with respect to such characteristics, called confounders, may lead to very misleading estimates of the predictive performance. Many strategies have been proposed to handle confounders, ranging from data selection, to normalization techniques, up to the use of training algorithm for learning with imbalanced data.
However, all these solutions require the confounders to be known a priori. To this aim, we introduce a novel index that is able to measure the confounding effect of a data attribute in a bias-agnostic way.  This index can be used to quantitatively compare the confounding effects of different variables and to inform correction methods such as normalization procedures or ad-hoc-prepared learning algorithms. The effectiveness of this index is validated on both simulated data and real-world neuroimaging data.

\end{abstract}

\end{frontmatter}

\section{Introduction}
In the last years, there has been a growing interest in the use of supervised learning in biomedical contexts.
However, such biomedical applications are often subject to the detrimental effects of so-called confounders, that are characteristics of the data generation process that do not represent clinically relevant aspects, but might nevertheless bias the training process of the predictor \cite{neto2018using, neto2017learning, greenland2001confounding}. In neuroimaging studies, for instance, the confounding effect of demographic characteristics such as gender and age is amply discussed \cite{rao2017predictive, brown2012adhd}. Studies on biometric sensor data, instead, have shown that the relationship between features and disease class label learned by the classifier is confounded by the identity of the subjects, because the easier task of subject identification replaces the harder task of disease recognition \cite{saeb2017need, neto2017learning}. Finally, learning algorithms trained with a collection of different databases, a common practice in biomedical applications, suffer from high generalization errors caused by the confounding effects of the different acquisition modalities or recruitment criteria \cite{zhao2018multiple}. This phenomenon is often referred to as 'batch effect' in gene-expression studies \cite{lazar2012batch} and it is proved that it may lead to spurious findings and hide real patterns \cite{scherer2009batch,lazar2012batch,leek2010tackling,akey2007design,parker2012practical,soneson2014batch}.\\

The acknowledgement of these problems brought us to a precise definition of a confounder as a variable that affects the features under examinations and has an association with the target variable in the training sample that differs from that in the population of interest \cite{rao2017predictive}. In other words, the training set contains a bias with respect to such confounding variable. The approaches developed to deal with confounders can be summarized in three broad classes. The first and most intuitive one matches training data with respect to the confounder, thus eliminating the bias, at the cost of discarding subjects and impoverishing the dataset \cite{Rao2017PredictiveMU,neto2018using}. A second approach corrects data with a normalization procedure, regressing out the contribution of the confounder before estimating the predictive model \cite{dukart2011age,abdulkadir2014reduction,Rao2017PredictiveMU}. However, the dependency of the data from the confounders may not be trivial to capture in a normalization function and this problem is exacerbated when different confounders are considered together. For example, batch effects cannot be easily eliminated by the most common between-sample normalization methods \cite{leek2010tackling, luo2010comparison}. Alternatively, confounders have been included as predictors along with the original input features during predictive modelling \cite{rao2015comparison,Rao2017PredictiveMU}. However, it has been noted that the inclusion in the input data of a confounder that is highly associated with the correct response may actually increase its effect, since in this case the confounder alone can be used to predict the response. Recently, a third kind of more articulated approaches has been developed, operating on the learning model rather than on the data itself, for instance resorting to Domain Adaptation techniques \cite{zhao2018multiple}. Similarly, some attempts have been made using approaches designed to enforce fairness requirements in learning algorithms, so that sensitive information such as ethnicity does not influence the outcome of a predictor \cite{hardt2016equality,zafar2017fairness,calders2009building,donini2018empirical}.
However, also in these models, it is very difficult to correct for multiple confounders, as it would be necessary in biomedical studies.

An effective solution to the confounders problem thus requires combining the three techniques described above: normalizing for the confounders that have a known effect on the data, matching the subjects if this does not excessively reduce sample size, and adopting a learning algorithm able to manage the biases that have not been eliminated earlier. When planning such an articulated approach, it is useful to have an instrument that can quantify the effect of a confounding variable and assess the effectiveness of the possible countermeasures. To this aim, we present in this paper a novel figure of merit, called 'Confounding Index' (CI) that measures the confounding effect of a variable in a binary classification task tackled through Machine Learning (ML) models.

Previous renowned works on this subject are the 'Back-door' and 'Front-door' criteria, developed in the causal inference framework described in Judea Pearl's work \cite{pearl1995causal,pearlcausal}, commonly cited as a way to determine which variables play as counfounders. However, both these criteria are not specifically developed for ML analysis and are based on conditional probabilities, thus, they provide a measure of the confounding effect that mainly depends on the specific composition of the dataset under examination. On the contrary, our CI is designed for ML problems and aims at quantifying how easily the way a confounder affects the data is learnable by the chosen algorithm with respect to the desired classification task, independently from the confounder distribution in the dataset.
Furthermore, given that the mentioned criteria do not take into account the algorithm used for the statistical analysis, they cannot be used to evaluate the effectiveness of an algorithm that, for example, has been specifically designed to avoid learning from biases.
To our knowledge, there is a single and recent study \cite{neto2018using} that, similarly to our purposes, presents a method of quantifying confounders effects in ML studies. However, this measure (thoroughly investigated in Section \ref{Elias}) is again strictly related to the specific biases present in the training set.\\

The proposed CI founds on measuring the variation of the area under the receiver operating characteristic curves (AUCs) obtained using different, engineered biases during training, and thus depends on how the confounder and the class labels affect the input features. The CI ranges from $0$ to $1$ and allows:
\begin{itemize}
\item to test the effect of a confounding variable on a specific binary classifier;
\item to rank variables with respect to their confounding effect;
\item to anticipate the effectiveness of a normalization procedure and assess the robustness of a training algorithm against confounding effects.
\end{itemize}
While the proposed approach is described for binary confounding variables, it can be applied for discrete ones computing the CI metric for every
pair of values and can be straightforwardly adopted also to assess the confounding effect of continuous variables by discretizing their values (an example of this is shown in the empirical assessment on our index). In such a scenario, CI allows to identify the widest range of values for which the effect of such variables can be ignored.\\

The biomedical sector is the one that we believe to be more suitable for the application of our CI since biomedical data, far more than other data types, depend in complex ways on many known and hidden factors of the data generation process.  However, the proposed CI is general enough to be applied in any supervised classification setup. The remainder of the paper is organized as follows: Section 2 introduces the formalization of the problem and the notation used throughout the paper, Section 3 discusses in detail the only other related work on this topic in literature. Section 4 and 5 describe the CI and its implementation.
Sections 6 and 7 report the experimental setup and the results of the analysis performed on both simulated and real-world neuroimaging data, while Section 8 concludes the paper.
A summary of the symbols used to describe the CI and the formulation of the confounding problem is reported in Table \ref{symbols}.
\definecolor{light-gray}{gray}{0.6}
\definecolor{lg}{gray}{0.9}

\begin{table}
\centering
 \resizebox{\textwidth}{!}{
 \begin{tabular}{|l | l|}
 \hline\hline\multicolumn{2}{|c|}{ }\\[-6pt]
 
 \multicolumn{2}{|c|}{\textbf{Notation to describe the confounding effect in a binary classification problem (Section \ref{notation}}).}\\[4pt] 

 \hline\hline&\\[-6pt]

 $\mathbf{\left(i, j\right)}$ & Sample and feature indexes. \\
 $\Vec{\mathbf{e}}$ & Canonical basis.\\[4pt]

 \arrayrulecolor{lg}\hline\arrayrulecolor{black}&\\[-6pt]
 
 $\mathbf{X}$ & L-dimensional space of the feature vectors $\Vec{\mathbf{x_i}}$.  \\
 
 $\mathbf{Y=\{+1,-1\}}$ & Space of the binary output labels $\mathbf{y_i}$. \\
 
 $\mathbf{C=\{\alpha,\beta\}}$ & Space of the confounding variable $\mathbf{c}$. \\
 
 $\mathbf{\digamma}$ & Space of the possible inference models $f$.\\[4pt]
 
 \arrayrulecolor{lg}\hline\arrayrulecolor{black}&\\[-6pt]
 
 $\mathbf{\Delta \subset  \left(X\times Y\right)}$ & Domain of the classification task.\\
 
 $\mathbf{D \subset \Delta}$ & Training set.\\
 
 $\mathbf{V \subset \Delta}$ & Validation set $\left(D \cap V = \varnothing \right)$.\\[4pt]
 
 \arrayrulecolor{lg}\hline\arrayrulecolor{black}&\\[-6pt]
 
 $\mathbf{\Delta^+}$ \; \small{\textcolor{light-gray}{$\left(\Delta^-\right)$}} & Subsample of $\Delta$ containing only elements with $y=+1$ \; \small{\textcolor{light-gray}{($y=-1$)}}.\\[1ex]
 
 $\mathbf{D^{+\alpha}}$ & Subsample of $D$ containing only elements with $y=+1$ and $c=\alpha$.\\
  \small{\textcolor{light-gray}{$\left(D^{-\alpha},D^{+\beta},D^{-\beta}\right)$}}&
  \small{\textcolor{light-gray}{(idem for the other combinations of values of $c$ and $y$)}} \\[1ex]

 $\mathbf{V^{+\alpha}}$ & Subsample of $V$ containing only elements with $y=+1$ and $c=\alpha$.\\
  \small{\textcolor{light-gray}{$\left(V^{-\alpha},V^{+\beta},V^{-\beta}\right)$}}&
  \small{\textcolor{light-gray}{(idem for the other combinations of values of $c$ and $y$)}}\\[4pt]
 
 \arrayrulecolor{lg}\hline\arrayrulecolor{black}&\\[-6pt]
 
 \multirow{2}{*}{$\mathbf{T \left(\cdot,\cdot\right)}$} & Training function, It takes the training sets of the the 2 classes\\
 &as inputs and returns a trained classifier function $f$.\\[1ex]
 
 \multirow{2}{*}{$\mathbf{AUC_f \left(\cdot,\cdot\right)}$} & Function that takes the validation sets of the the 2 classes\\
 &as inputs and returns the $AUC$ of the classifier $f$ on these sets.\\[4pt]
 
 \arrayrulecolor{lg}\hline\arrayrulecolor{black}&\\[-6pt]

 \multirow{2}{*}{$\mathbf{z_{ij}}$} & Scalar weight that depends on the characteristic of\\
 &the example $i$ and on the type of feature $j$.\\
 
 $\mathbf{g_{+}}$ & Function that describes how the feature vectors depend from $y = +1$.\\
 \small{\textcolor{light-gray}{$\left(g_{-},g_{\alpha},g_{\beta}\right)$}}& 
 \small{\textcolor{light-gray}{(idem for the other values of $c$ and $y$)}} \\[1ex]
 
 $\mathbf{I^{+}}$ & List of the feature vector positions that depend on $y = +1$.\\
 \small{\textcolor{light-gray}{$\left(I^{-},I^{\alpha},I^{\beta}\right)$}}& 
 \small{\textcolor{light-gray}{(idem for the other values of $c$ and $y$)}} \\[4pt]
  
  \hline\hline\multicolumn{2}{|c|}{ }\\[-6pt]
  
  \multicolumn{2}{|c|}{\textbf{Notation used to describe the proposed $CI$ (Section \ref{def_CI}).}}\\[4pt] 

 \hline\hline&\\[-6pt]
  
 $\mathbf{b}$ & Percentage bias between variables $c$ and $y$ in the training set.\\
 $\mathbf{f_b}$ & Classifier function obtained from a training set with a bias $b$.\\[4pt]
 
 \arrayrulecolor{lg}\hline\arrayrulecolor{black}&\\[-6pt]
 
 $\mathbf{D^{+\alpha}_N}$ & $D^{+\alpha}$ with size $N$.\\
  \small{\textcolor{light-gray}{$\left(D^{-\alpha}_N,D^{+\beta}_N,D^{-\beta}_N\right)$}}&
  \small{\textcolor{light-gray}{(idem for the other combinations of values of $c$ and $y$)}} \\[4pt]
  
  \arrayrulecolor{lg}\hline\arrayrulecolor{black}&\\[-6pt]
  
 \multirow{3}{*}{$\mathbf{\Phi_{pro}}$ \; \small{\textcolor{light-gray}{($\Phi_{cons}$)}}} & 
 Measure that represents the increase \small{\textcolor{light-gray}{(decrease)}} with respect to\\ &$AUC_{f_{b=0}}$ in case the training and validation sets are biased in the\\
 &same \small{\textcolor{light-gray}{(opposite)}} way with respect to the value of $c$.\\[1ex]
 
 \multirow{2}{*}{$\mathbf{\Phi=\Phi_{cons} +\Phi_{pro}}$}  & 
 Figure of merit that summarizes the effects that various degrees of $b$\\ &in the training set with respect to a particular classification problem. \\[1ex]
 
 \multirow{4}{*}{$\mathbf{\Phi^*}$}  & 
 Measure equivalent to $\Phi$, but obtained from a different training configuration.\\& If $\Phi$ is calculated considering a training with a positive correlation between\\& the pairs $(y=+1,c=\beta)$ and $(y=-1,c=\alpha)$, $\Phi^*$ considers a positive correlation\\&  between the pairs $(y=+1,c=\alpha)$ and $(y=-1,c=\beta)$.\\[1ex]
 
 $\mathbf{CI = \max({\Phi, \Phi^*})}$ & The proposed CI index.\\[4pt]
 
 \hline

 \end{tabular}}
 \caption{Summary of the symbols introduced in Sections \ref{notation} and \ref{def_CI} to describe the problem of confounders in a binary classification framework and to define our $CI$.}
 \label{symbols}
\end{table}

\section{Notation}\label{notation}
In this section we introduce the notation used in this paper to describe a binary classification framework and the problem of confounders we want to address.\\
\subsection{Two-class machine learning}\label{two_class_notation}
In a typical two-class classification task, we define the L-dimensional input feature vectors  $\vec{x_i}\in X$ and their binary output labels $y_i \in Y = \{+1,-1\}$, where $i$ is the sample index.
The domain $\Delta$ of the task can be defined as a subset of $X \times Y$:
\begin{equation}
    \Delta = \big\{(\vec{x}_{1}, y_1),(\vec{x}_{2}, y_2), ...,(\vec{x}_{n}, y_n)\big\}.
\end{equation}
We suppose that the feature vectors depend on their output label and other unknown variables, which can be considered irrelevant for the applicability and effectiveness of the classification algorithm.\\
Let us define two subsamples of $\Delta$ called $\Delta^+$ and $\Delta^-$ as follows:
\begin{equation}
    \begin{array}{l}
    \Delta^+ = \big\{(\vec{x}_{i}, y_i) \; : \; y_i = +1\big\}\\
    \Delta^- = \big\{(\vec{x}_{i}, y_i) \; : \; y_i = -1\big\}
    \end{array}.
\end{equation}
From here on we will refer generically to any of these subsamples using $\Delta^\pm$.\\
Training a two-class ML algorithm can be represented as a function $T$ as follows:
\begin{equation}
    T\;:\mathcal{P}(\Delta^+)\times \mathcal{P}(\Delta^-) \rightarrow \digamma ,
\end{equation}
where $\digamma$ is the space of the possible inference models $f$ such that $f:X \rightarrow Y$ and $\mathcal{P}(\Delta^\pm)$ denotes the power set of $\Delta^\pm$, i.e., the set of all possible subsets of $\Delta^\pm$. Thus, training a binary classifier can be viewed as finding the function $f=T(D^+,D^-)$, where $D^+\in\mathcal{P}(\Delta^+)$ and $D^-\in\mathcal{P}(\Delta^-)$.\\
The function $f$ is chosen from the subset of the inference models that can be explored by the chosen algorithm as the one that minimizes the error made in labelling only the elements $\vec{x_i}$ of the training set $D = D^+ \cup D^-$.\\
To evaluate the generalizability of $f$, various figures of merit exist that quantify the error resulting from the application of $f$ on samples external to the training set. The commonest ones are Accuracy, Sensitivity, Specificity and AUC. 
Between these quantities, the AUC is the only one that takes into account all the four possible outcomes of the classifiers (true positive, false positive, true negative and false negative).\\
It is calculated as the area under the curve obtained by plotting the true positive rate against the false positive rate at various discrimination threshold settings. Thus, the AUC is a measure of the classification performances of a classifier that is independent on the specific threshold used to assign the labels and it is commonly considered the most reliable metric for the evaluation of machine learning algorithms \cite{bradley1997use}.\\
For this reason, both our work and the one described in \cite{neto2018using}, are based on this quantity. In particular, for the definition of the CI we need to introduce $AUC_f(V^+,V^-)$ as a function that returns the AUC of a model $f$ on a validation set $V=V^+ \cup V^-$, a subset of $\Delta$ without any intersections with the training set $D$.

\subsection{The confounding effect}
Let us suppose that among the various unknown variables that affect the values of $\vec{x_i}$, there is a variable $c$ (with no causative effect on the actual class label $y$ and with values in $C = \{ \alpha,\beta \}$) that might have a confounding effect on the classification algorithm, while the others remain irrelevant for our purpose. In this situation we consider $\vec{x_i}$ as mainly dependent on two variables $c$ and $y$ and thus we can define four subsamples of $X$: $X^{+\alpha}$, $X^{-\alpha}$, $X^{+\beta}$ and $X^{-\beta}$.\\
For instance, $X^{+\alpha}$ is defined as follows:
\begin{equation}
    X^{+\alpha} =\big\{ \vec{x}_{i}(c_i,y_i) \; : \; y_i = +1, c_i = \alpha \big\}
\end{equation}
and similarly for the others.
Every vector of features can be described as the sum of three contributions. One term represents the dependence from the unknown variables, while the other two represent the dependence from the variables $y$ and $c$. For example, $\vec{x_i} \in X^{+\alpha}$ (and similarly for the other subsamples) can be written as:
 \begin{equation}\label{input_data_description}
   \vec{x_i} \in X^{+\alpha} \; : \;\; \vec{x_i} = 
    \sum_{j=1}^{L} z_{ij} \cdot e_j + 
    \sum_{j \in I^{\alpha}} g_{\alpha}(z_{ij},j) \cdot e_j +
    \sum_{j \in I^+} g_{+}(z_{ij},j) \cdot e_j
\end{equation}
Where $L$ is the dimension of the feature vectors $\vec{x_i}$, $\vec{e}$ is the canonical basis and $z_{ij}$ are scalar weights that depend on the characteristic of the specific example $i$ and on the type of feature $j$. 
$I^{\alpha}$ and $I^+$ are the feature vector positions that depend, respectively, on the variable $c=\alpha$ and on the label variable $y=+1$. 
The terms $x_{ij}$, with $j\in I^{\alpha}$, depend on the function $g_{\alpha}$, while those with $j\in I^{+}$ depend on $g_{+}$.
These functions describe how the the values of $c$ and $y$ affect the feature vectors.\\
The main objective of a binary classifier trained to distinguish between $X^+$ and $X^-$ is to learn a pattern based on the differences introduced in $\vec{x_i}$ by $g_{+}$ and $g_{-}$.
However, when the patterns due to $g_\alpha$ and $g_\beta$ are more easily identifiable than the ones of interest, and the distribution of the values of $c$ is uneven across the training samples $D^+$ and $D^-$, the classifier can be misled and we say that $c$ has a confounding effect on the classification task.

\section{Related Works}\label{Elias}
Previous literature on the effect of confounders in ML analysis is divided mainly on statements of the problem and solution proposals (i.e. normalization procedures, corrected loss-functions, etc.). However, every study estimates the confounding effects on its results differently, often taking arbitrary decisions about which possible confounders to consider and how.\\
Our objective is thus to propose our CI as a standardized tool to quantify the effect of a variable that may play as confounder in an analysis based on a specific classifier and with a specific classification task.\\
In this section we briefly illustrate the only work \cite{neto2018using} in literature with a similar aim, which present a permutation based confounding estimate.
Through a simulation study, we discuss its limits and its differences with respect to our proposal.
For consistency, we adopt the notation previously described.\\

\subsection{Permutation based Confounding Estimate}
\begin{figure}
 \centering
  \includegraphics[scale=0.65]{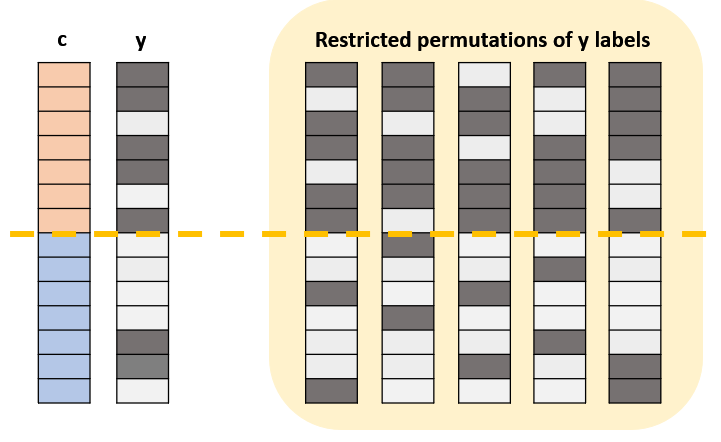}
  \caption{Graphical representation of the restricted permutation. The two columns on the left show the distribution of the $c$ and $y$ values in the sample. Red and blue cells represent observations with respectively $c=\alpha$ and $c=\beta$. Dark and light gray cells represent observations with respectively $y=+1$ and $y=-1$. The orange dashed line divides the items based on their values of $c$. In the light orange box there are some examples of how to assign the labels $y$ in a restricted permutation test. Basically, the $y$ labels are shuffled maintaining the same percentage of observations in the groups defined by the same value of pair $(c,y)$.}
  \label{restricted}
\end{figure}

\begin{figure}
 \centering
  \includegraphics[scale=0.55]{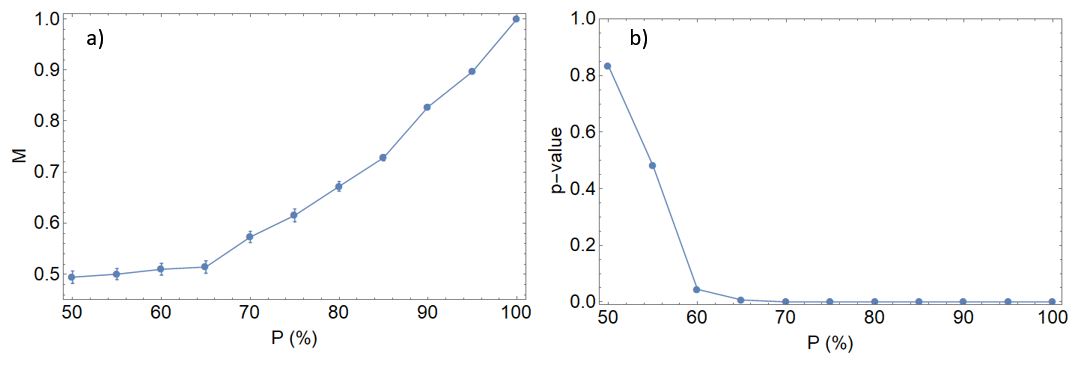}
  \caption{Plot showing the values of $M$s (Fig. a) and of the p-values (Fig. b) obtained without any confounding effect related to the variable $c$, with respect to the percentage $P$ of bias between $y$ and $c$.}\label{M}
\end{figure}

In the context of predictive modeling, the authors of the work \cite{neto2018using} propose to use a Restricted Permutation (RP) to isolate the contribution of a categorical confounder $c$ from the performance of a binary classifier trained to predict the class label $y$.
Fig. \ref{restricted} shows the RP procedure, in which basically the class labels are shuffled separately for each value of $c$.\\
The authors claim that this permutation "destroys the direct association between the response and the features while still preserving the indirect association due to the confounder".
The Standard Permutation (SP), that randomly shuffles the labels without any restriction, destroys instead both the the association between $y$ and the features and between $y$ and $c$.\\
From these considerations, they sustain that the distribution of the $AUCs$ obtained training a classifier on datasets generated with RP is equivalent to the one obtained with SP under the null hypothesis that an algorithm has not learned the confounding signal. In the presence of confoundings, instead, the RP distribution will be shifted away from the SP.\\
Thus $M$, the average $AUC$ of the RP distribution, is defined by the authors as a 'natural measure of the amount of confounding signal learned by the algorithm'.
A statistical test to evaluate whether or not the algorithm has learned the confounding signal is naturally represented by a p-value that verifies if $M$ belongs to the SP null distribution.
Considering that this last distribution, according to the authors, is known to be approximated by a normal distribution centered in 0.5 and with a variance that depends on the test set composition \cite{bamber1975area,mason2002areas}, the p-value they propose requires to compute only $M$.\\
Our concerns arise from the fact that, considering how the restricted permutation is implemented, it does not completely remove the dependence of the response $y$ from the data (as expected by the authors). For example, let us consider a bias of 95\%, which means that 95\% of the data with $y=1$ has $c=\alpha$. In this situation, the restricted permutation shuffles the y labels maintaining the same proportion of data with $(y=1,c=\alpha)$, and thus, will assign the label y=1 to the data truly belonging to class 1 with a higher probability. Therefore, the algorithm can learn the response signal even when the dataset is shuffled with the RP, leading to an overestimation of the confounding effect.\\
We tested this hypothesis with simulated data consisting in two-feature vectors in which the variable $y$ affects the first feature, while the variable $c$ affects the second one.
In our experiment $c=\alpha$ and $c=\beta$ have the same effect on the feature vectors, because we want to test whether the restricted permutation test is able to correctly identify $c$ as not confounding.\\
In Fig. \ref{M}a we report the values of $M$ as a function of the percentage $P$ of input data with $(y,c)=(+1,\alpha)$, which represents the bias in the training phase. 
If the assumptions made by the authors were correct we should have obtained $M \approx 0.5$ independently from the value of $P$, because in these simulated data $c$ is not confounding. However, as shown in Fig. \ref{M}a, $M$ increases monotonically with respect to the value of the bias $P$. Given that these data depend only on $c$ and $y$, but are not differentiated with respect to the values of $c$, the increase of $M$ can be attributed only to the dependence of the data from $y$. Fig. \ref{M}b shows the p-value obtained from the $M$ values in Fig. \ref{M}a, as it can be noted the statistical test makes an error of type 1 when $y$ and $c$ are sufficiently correlated. 

\subsection{Differences with the proposed CI}
With the previous discussion, we have shown that the RP method consistently overestimates the confounding effect of a variable because it is unable to eliminate the association between class label $y$ and the features.
We believe this is due to the assumption that the RP distribution is equivalent to the SP distribution, under the null hypothesis, which is valid only when the null hypothesis means that there is no correlation between $y$ and $c$, without considering the null hypothesis in which $y$ and $c$ might be correlated, but $c$ does not affect the data.
This is incompatible with the aim of our $CI$, that is intended to be used in a preliminary study to understand which data attributes may mislead the learning process.
Furthermore, another intrinsic limit of this permutation based approach is that it measures a quantity that depends on the specific bias present in the training set. This makes it difficult to rank the effect of different confounders.

\section{Definition of Confounding Index (CI)} \label{def_CI}
In this section, we present our definition of Confounding Index (CI). This index makes it possible to compare the confounding effects of categorical variables with respect to a defined two-class classification task, with a measure that does not depend on the particular bias present in the training dataset.\\
Basically, it shows how easily the differences due to a possible confounder can be detected by a ML algorithm with respect to the differences due to the classes we want to study. The applicability of the proposed CI can be extended also to study the confounding effects of continuous variables with an appropriate binning.
We begin by defining and discussing the validity of our CI. Then, we provide a pseudocode description for computing it.

\subsection{CI definition}\label{def_subsec}
In order for our CI not to depend on the particular bias of the training set, we have to study how the model function $f$ obtained with the training varies with respect to the bias $b$.
Thus, let us consider a group of model functions obtained using different compositions of the training set (in the following notation the set subscripts denote the size of the samples):
\begin{equation}\label{f_b_training}
     f_b = T\Big(D^{+\alpha}_{\scaleto{N(1-b)}{4.5pt}} \cup D^{+\beta}_{\scaleto{N(1+b)}{4.5pt}}\;, \;\;
                D^{-\alpha}_{\scaleto{N(1+b)}{4.5pt}} \cup D^{-\beta}_{\scaleto{N(1-b)}{4.5pt}}\Big) \;\;\; where\;\; b=\frac{a}{N},\; 0 \leq a \leq N, a \in \mathbb{N}.
\end{equation}
When $b=0$, training does not present any bias with respect to the confounding variable, thus, the error committed by the model $f_0$ should not depend on the distribution of $c$ over the validation set, which means that all the following expressions should be equivalent (except for finite sample effects):

\begin{equation}\label{ugualianza_validazioni}
\begin{aligned}
AUC_{f_0}\Big(V^{+\alpha} \cup V^{+\beta}\;, \;V^{-\alpha} \cup V^{-\beta}\Big) &= AUC_{f_0}\Big(V^{+\alpha}\;, \;V^{-\beta}\Big) \\ &= AUC_{f_0}\Big(V^{+\beta}\;, \;V^{-\alpha}\Big)
\end{aligned}
\end{equation}

Thus, from now on, we will use the term $AUC_{f_0}$ to refer to any of these three values.\\
For $b\neq0$, $f_b$ is obtained from a biased training with bias $b$, which means that $f_b = f_0$ only if $g_{\alpha} = g_{\beta}$ and $I^{\alpha} = I^{\beta}$ are both true (as above except for finite sample effects). In this case, Eq. (\ref{ugualianza_validazioni}) should hold for a generic $f_b$, too.
From this observation it derives that the following condition is necessary for $c$ to be a confounding variable:
\begin{equation}\label{confounding_nec1}
    I)\;\;\; \exists \; b' \neq 0 : \;\;\;\;
    AUC_{f_{b'}}\Big(V^{+\alpha}\;, \;V^{-\beta}\Big) \neq AUC_{f_{b'}}\Big(V^{+\beta}\;, \;V^{-\alpha}\Big).
\end{equation}
In particular, considering which samples are more and less represented in Eq. (\ref{f_b_training}), if $c$ is confounding enough to affect the training phase, we can expect a monotone increase in $AUC_{f_b}\big(V^{+\beta}\;, \;V^{-\alpha}\big)$  with respect to $b$. This happens because the training and validation sets are biased in the same way with respect to the values of $c$.
Conversely, when the training and validation sets are oppositely biased, the logic that the model learns during the training phase no longer holds in the validation phase and thus the value $AUC_{f_b}\big(V^{+\alpha}\;, \;V^{-\beta}\big)$ should monotonically decrease. Summarizing, if the variable $c$ has a confounding effect both these monotonicity conditions should hold:
\begin{equation}\label{confounding_nec2}
 \begin{split}
    II)\;\;\; AUC_{f_b'}\Big(V^{+\alpha}\;, \;V^{-\beta}\Big) &\geq AUC_{f_b''}\Big(V^{+\alpha}\;, \;V^{-\beta}\Big) \;\; \forall b'<b'' \\
    III)\;\;\; AUC_{f_b'}\Big(V^{+\beta}\;, \;V^{-\alpha}\Big) &\leq AUC_{f_b''}\Big(V^{+\beta}\;, \;V^{-\alpha}\Big) \;\; \forall b'<b''. \\
 \end{split}
\end{equation}
When all the three conditions in Eq. (\ref{confounding_nec1}) and (\ref{confounding_nec2}) are satisfied, we define $c$ as a confounder and we can formulate a possible figure of merit to quantify its effects as the difference between the two terms of inequality (\ref{confounding_nec1}), integrated over $b$:
\begin{equation}\label{C}
    \Phi = \int_{0}^{1} \Big[AUC_{f_b}\Big(V^{+\beta}\;, \;V^{-\alpha}\Big) - AUC_{f_b}\Big(V^{+\alpha}\;, \;V^{-\beta}\Big)\Big]  \; db.
\end{equation}
Eq. (\ref{C}) can be rewritten in a more explicit form as the sum of two contributions $\Phi_{cons}$ and $\Phi_{pro}$ which are:
\begin{equation}\label{CI_explained}
\begin{split}
    \Phi_{cons} &= \int_{0}^{1}\Big[ AUC_{f_0} - AUC_{f_b}\Big(V^{+\alpha}\;, \;V^{-\beta}\Big) \Big]\;db \\
    \Phi_{pro} &= \int_{0}^{1}\Big[ AUC_{f_b}\Big(V^{+\beta}\;, \;V^{-\alpha}\Big) - AUC_{f_0} \Big]\;db,
 \end{split}
\end{equation}
where $\Phi_{pro}$ ($\Phi_{cons}$) represents the increase (decrease) with respect to $AUC_{f_0}$ in case the training and validation sets are biased in the same (opposite) way with respect to the value of $c$.
It is important to consider both these terms because, since $AUC$ is confined in the interval $[0,1]$, depending on the value of $AUC_{f_0}$, one of the two contributions may not correctly reflect the effect of the bias.\\
The figure of merit just described, $\Phi$, summarizes the effects that various degrees of bias in the training sets have in a particular classification problem.
However, it has been defined studying the $AUCs$ of the group of model functions described in Eq. (\ref{f_b_training}), that have been constructed using a positive correlation (measured by the bias $b$) between the pairs of labels $(y=+1,c=\beta)$ and $(y=-1,c=\alpha)$, while completely neglecting the possibility of the inverse situation: a positive correlation between the pairs $(y=+1,c=\alpha)$ and $(y=-1,c=\beta)$.
Given that we want to define a measure that does not depend on the particular bias present in the training dataset, we should consider also the $\Phi^*$ calculated from the group of model functions $f^*_b$:
\begin{equation}\label{C^*}
 \begin{split}
 f^*_b &= T\Big(D^{+\beta}_{\scaleto{N(1-b)}{4.5pt}} \cup D^{+\alpha}_{\scaleto{N(1+b)}{4.5pt}}\;, \;\;
                D^{-\beta}_{\scaleto{N(1+b)}{4.5pt}} \cup D^{-\alpha}_{\scaleto{N(1-b)}{4.5pt}}\Big)\\
 \Phi^* &= \int_{0}^{1} \Big[AUC_{f^*_b}\Big(V^{+\alpha}\;, \;V^{-\beta}\Big) - AUC_{f^*_b}\Big(V^{+\beta}\;, \;V^{-\alpha}\Big)\Big]  \; db.
 \end{split}
\end{equation}
To understand why $\Phi$ and $\Phi^*$ can differ, let us consider for example the case in which both the differences due to $y$ and $c$ are not easily understandable by the model, and $I^\beta = I^+$ while $I^\alpha \neq \{I^+, I^-\}$. In this situation the effects of a bias can depend on which correlation we choose when building the training set, and as a consequence also the validation set. This happens because when we compute $\Phi$ we are measuring the effects of a positive correlation between the pairs $(y=+1,c=\beta)$; given that both these variables are affecting the same features, the confounding effect will be higher with respect to the case of the opposite correlation, measured with $\Phi^*$.\\
Given that the real correlation between these two variables is unknown and that we want to measure the confounding effects in the worst case possible, we define our CI as:
\begin{equation}\label{CI_final}
    CI = \max({\Phi, \Phi^*}).
\end{equation}

\subsection{CI applicability}\label{applicability}
Looking at the definitions of $\Phi$ and $\Phi^*$ in Eq. \ref{C} and \ref{C^*} it is clear that both these measurements have values in $[-1,1]$, because they represent a difference between two numbers, the $AUCs$, constrained to be in $[0,1]$ and integrated over a range of unitary length.\\
However, as previously stated, $\Phi$ (and thus also $\Phi^*$), can be used to quantify the confounding effects of $c$ only if the conditions in Eq. (\ref{confounding_nec1}) and (\ref{confounding_nec2}) are valid. 
In particular, if the monotonicity conditions hold, $\Phi$ and $\Phi^*$ have values in $[0,1]$ (because the integrand is positive), thus this is also the range of meaningful values of the proposed $CI$, where 1 indicates the maximum confounding effect measurable, while 0 means the absence of any effect.
The monotonicity evaluation can be performed both using automated techniques or even with just a visual inspection of the data.\\
Dealing with real data, which can be scarce and noisy, the values of the $AUCs$ can oscillate significantly. We therefore suggest to repeat the $AUCs$ calculation more than once, adopting the desired resampling method (e.g. bootstrap, cross validation, etc.) and using the mean values to compute the proposed $CI$. 
There are no preferred resampling methods because the value of $CI$ is computed as the sum of $\Phi_{pro}$ and $\Phi_{cons}$ which represent respectively the increase and decrease with respect to $AUC_{f_0}$. Thus, if for instance one resampling method systematically underestimate the performance of the $AUCs$,  this will result in a lower $\Phi_{pro}$ and and a greater $\Phi_{cons}$ and their sum will not be affected by this underestimation. During the computation of the $CI$ what really matters is to be consistent with the resampling choice adopted and to properly evaluate the propagation of the errors associated to the estimates.\\
Another fact to take into account is that the $CI$ has been defined under the hypothesis that the input data depend mainly on two variables: the label one $y$ and a possibly confounding one $c$, and considering the dependency from other variables irrelevant for the classification purposes. If this condition is not valid and the data depend strongly on other variables, it is very important to match the training sample with respect to these ones. Skipping this match operation may introduce unwanted biases in the computed $AUCs$ with respect to these confounding variables, causing a violation of the monotonicity conditions. 
\\
Summarizing, the steps to correctly evaluate the confounding effect of a variable $c$ are: 
\begin{itemize}
    \item Build the various training and validation datasets necessary for the $CI$ calculation, matching the data for other possible variables that can affect this analysis.
    \item Compute the $AUCs$ needed to evaluate the confounding effect of $c$ and use their average values for the calculation of $\Phi$ and $\Phi^*$.
    \item Evaluate the monotonicity conditions.
    \item Between $\Phi$ and $\Phi^*$, take the one that satisfies the monotonicity conditions as the value of $CI$. If both of them do, take the largest one.
    If none of them does, the CI is undefined: probably the data have not been correctly matched for one or more confounding variables.
\end{itemize}

\subsection{Pseudo-code for the calculation of the CI}
In this section, we describe the two algorithms necessary to evaluate the confounding effect of a variable $c$. \\
Algorithm \ref{ccomp} calculates the quantity $\Phi$ (the calculation of $\Phi^*$ is equivalent). The algorithm begins by initializing two lists: $AUC^{pro}$ and $AUC^{cons}$. The first (second) one will be filled with the performance metrics computed, for the different $b$ explored, when training and validation sets are biased in the same (opposite) way with respect to the values of $c$. For the sake of clarity, $AUC^{pro}$ and $AUC^{cons}$ will collect respectively the estimates of the various $AUC_{f_b}(V^{+\beta},V^{-\alpha})$ and $AUC_{f_b}(V^{+\alpha},V^{-\beta})$ in Eq. (\ref{C}).
The algorithm initially computes $AUC_{f_0}$ on an unbiased dataset and puts it into the two lists. Then, for each bias explored, several different dataset compositions are used to compute the aforementioned $AUCs$ and the results are averaged together to get more robust estimates of the values $AUC_{f_b}(V^{+\beta},V^{-\alpha})$ and $AUC_{f_b}(V^{+\alpha},V^{-\beta})$. These estimates are then appended to the respective lists. $\Phi$ is computed as the difference between the areas under the curves drawn by the elements in the lists $AUC^{pro}$ and $AUC^{cons}$ respectively.
With respect to the theoretical description made in Section \ref{def_subsec}, in this algorithm, we correct $\Phi$ for the effect of the discrete steps used to explore the range of $b$. In fact, the maximum value of $\Phi$ for a finite step size is not 1 but $(1-step/2)$. Correcting for this factor is necessary in order to obtain a $CI$ value that can be easily compared to other ones obtained with a different step size.
Algorithm \ref{ccomp} ends with the assessment of the monotonicity conditions (Eq \ref{confounding_nec2}) and returns $\Phi$ and the results of these checks.
Note that, in order to get the most faithful results possible, during the computation of this algorithm, all training and validation sets should be carefully matched for all the possible confounding variables that are not under study.\\
Finally, algorithm \ref{cicomp} describes how to asses the $CI$ value. First, the values of $\Phi$ and $\Phi^*$ are computed as described in Algorithm \ref{ccomp}. Then, considering the values of $\Phi$ and $\Phi^*$ and whether they satisfy the monotonicity conditions, the correct CI value is returned.

\begin{algorithm}
\setstretch{1.5}
\caption{$\Phi$ computation}\label{ccomp}
\begin{algorithmic}[1]
\State {Let $4N$ be the total number of data points available}
\State {Initialize empty lists $AUC^{pro}$ and $AUC^{cons}$}
\State {Train the model on an unbiased dataset $D_{0} = D^{+\alpha}_{N} \cup D^{+\beta}_{N} \cup D^{-\alpha}_{N} \cup D^{-\beta}_{N}$ }
\State {Calculate $AUC_{f_0}$}
\State {Append $(0,AUC_{f_0})$ to $AUC^{pro}$ and to $AUC^{cons}$}

\State {Choose a step size $s\in \field{N} : 1 \leq s \leq N $}
\State {Choose the number of averages $M$ to compute.}
\For{$(b=s/N;\; b\leq1;\; b=b+s/N)$}:
    \State {Initialize empty list $AUC^{pro,b}$ and $AUC^{cons,b}$}
    \For{$(m=0;\; m<M;\; m=m+1)$}:
        \State {Build the training set:
        $D_{b} = D^{+\alpha}_{\scaleto{N(1-b)}{4.5pt}} \cup D^{+\beta}_{\scaleto{N(1+b)}{4.5pt}} \cup D^{-\alpha}_{\scaleto{N(1+b)}{4.5pt}} \cup D^{-\beta}_{\scaleto{N(1-b)}{4.5pt}}$
        }
        \State {Build the validation subsets:
        $V^{+\alpha}, V^{+\beta}, V^{-\alpha}, V^{-\beta}$
        }
        \State{Train the model on $D_b$ to obtain a classifier function $f_{b}$}
        \State{Calculate $AUC^{pro,m}_{f_{b}}$ on $V^{pro} = V^{+\beta} \cup V^{-\alpha}$}
        \State{Append $AUC^{pro,m}_{f_{b}}$ to the list $AUC^{pro,b}$}
        \State{Calculate $AUC^{cons,m}_{f_{b}}$ on $V^{cons} = V^{+\alpha} \cup V^{-\beta}$}
        \State{Append $AUC^{cons,m}_{f_{b}}$ to the list $AUC^{cons,b}$}
    \EndFor
    \State{Compute $\overline{AUC^{pro,b}}$, the average of $AUC^{pro,b}$} 
    \State{Append $(b,\overline{AUC^{pro,b}})$ to $AUC^{pro}$}
    \State{Compute $\overline{AUC^{cons,b}}$, the average of $AUC^{cons,b}$} 
    \State{Append $(b,\overline{AUC^{cons,b}})$ to $AUC^{cons}$}
    \State{\textbf{end}}
\EndFor
\State{\textbf{end}}
\algstore{myalg}
\end{algorithmic}
\end{algorithm}

\begin{algorithm}                 
\begin{algorithmic} [1]                   
\algrestore{myalg}
\setstretch{1.5}
\State{Compute $Pro$ as the area under the curve drawn by the list $AUC^{pro}$}
\State{Compute $Cons$ as the area under the curve drawn by the list $AUC^{cons}$}
\State{Compute $\Phi = Pro - Cons$}
\State{Correct for the step length $\Phi = \Phi/(1-\frac{s}{2N})$}
\State{Assess that the values in the lists $AUC^{pro}$ and $AUC^{cons}$ are monotonically increasing for the first and decreasing for the second.}
\State{Return both $\Phi$ and the result of the monotonicity assessment}
\end{algorithmic}
\end{algorithm}

\begin{algorithm}
\setstretch{1.5}
\caption{CI computation.}\label{cicomp}
\begin{algorithmic}[2]
\State{Compute $\Phi$ as detailed in Algorithm \ref{ccomp}}
\State{Compute $\Phi^*$ analogously}
\State{Four scenarios are possible:}
\Indent
    \State{1. Both $\Phi$ and $\Phi^*$ respect the monotonicity conditions}
    \Indent
        \State{Return $CI = \max{\{\Phi,\Phi^*\}}$}
    \EndIndent
    \State{2. Only $\Phi$ respects the monotonicity conditions}
    \Indent
        \State{Return $CI = \Phi$}
    \EndIndent
    \State{3. Only $\Phi^*$ respects the monotonicity conditions}
    \Indent
        \State{Return $CI = \Phi^*$}
    \EndIndent
    \State{4. Both $\Phi$ and $\Phi^*$ do not satisfy the monotonicity conditions}
    \Indent
        \State{$CI$ is undefined}
    \EndIndent
\EndIndent
\end{algorithmic}
\end{algorithm}
\section{Monotonicity evaluation}
As already explained in Section \ref{applicability}, our CI can be calculated only under the monotonicity conditions of Eq. (\ref{confounding_nec2}) which should be verified.
This can be done with just a visual inspection of the data, or using various trend analysis methods already described in literature.
In this section we will briefly illustrate the method presented in \cite{brooks2005scale} that we have used for all the analysis described in this paper.
We chose this method because it allows to evaluate the monotonicity conditions even when the trends may present noise and fluctuations. This may happen especially when computing the $CI$ on real data. In this case the monotonicity conditions should take into account the expected variability introduced by noise, finite-sample effects and the resampling method adopted to compute the various $AUC_{f_b}$.

\subsection{Delta-monotonicity}
\begin{figure}
 \centering
  \includegraphics[scale=0.65]{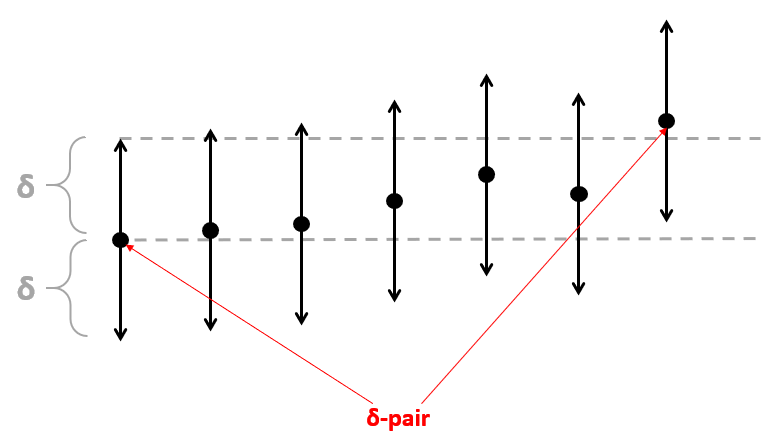}
  \caption{Example of $\delta-pair$, as described in \cite{brooks2005scale}}
  \label{delta_pair}
\end{figure}

The method described in \cite{brooks2005scale} is based on the notion of scale-based monotonicity, which means that the fluctuations within a certain scale are ignored.
The scale of the fluctuations can be chosen by the user and is called $\delta$.
Given a function $F$ defined over an ordered set of real values $D=\{x_1,x_2,...,x_n\}$, two elements $\{x_i,x_j\}$ where $j>i$ are called a $\delta-pair$ if their images under $F$ are significantly different, while the images of the points between them can be considered constant on the scale of $\delta$. This can be summarized with the following two conditions, graphically illustrated in Fig. \ref{delta_pair}:
\begin{itemize}
    \item $| F(x_j) - F(x_i) | \geq \delta$
    \item $\forall k \in D, i<k<j$  implies $| F(x_k) - F(x_i) | < \delta$ and $| F(x_k) - F(x_j) | < \delta$  
\end{itemize}
A $\delta-pair's$ direction is increasing or decreasing according to whether $F(x_j)>F(x_i)$ or $F(x_j)<F(x_i)$.
Given these preliminary definitions, the authors of \cite{brooks2005scale} define $F$ as $\delta-monotone$ over $D$ if all the $\delta-pairs$ have the same direction.

\section{Materials and Methods}
In this section we will first show the effectiveness of our $CI$ on simulated data and describe a possible application on real world data.\\
Artificial data in fact allow to analyze how $CI$ varies with respect to the differences introduced in the input data due to $c$ and $y$, while real world data can give a practical idea of the usefulness of the CI.\\
The real data used in this study are neuroimaging data \cite{di2014autism,di2017enhancing} which, as all the biomedical data, depend on several variables that can have a confounding effect on many different classification tasks. We will show that our index is able to rank the most important confounding variables that can affect a classifier, giving the possibility to design strategic solutions to the problem. We will also show that this index can be used for continuous variables, and in this case it can help in identifying a range of values in which the confounding effect of a variable can be considered irrelevant.\\
In all the analyses described in this section we have used a logistic regression classifier, but the CI can be constructed for any binary classifier.

\subsection{Simulated Data}
As already described in Section \ref{def_subsec}, we are considering the situation in which the input data $\vec{x_i} \in X$ of a classification problem depend mainly on two binary variables: the class label $y$ and a second variable $c$ that can have a confounding effect on the classification task. To assess the validity and effectiveness of our $CI$, we have generated four subgroups of artificial input data: $X^{+\alpha}$, $X^{-\alpha}$, $X^{+\beta}$ and $X^{-\beta}$. \\
The data belonging to every subgroup have been generated as explained in Eq. (\ref{input_data_description}).
In our simulation every $\vec{x_i}$ is a vector of 100 features, in which the first contribution (the one that was described as the linear combination of the $z_{ij}$ elements) has been simulated attributing to every feature a random real value in the range $[-10,10]$. The second and the third contributions (the ones that explain how the input data depend on $y$ and $c$) have been simulated adding or subtracting a constant value to a limited set of features, making the sum of all the contributions equal to zero.\\
In particular, to avoid the classifier learning a pattern based on the total sum of all the features, instead of finding which features are influenced by the value of a particular variable, the functions $g_{\alpha}$, $g_{\beta}$, $g_{+}$ and $g_{-}$ add a constant to two features and subtract the same constant from other two features.\\
We have tested our $CI$ in two different kinds of situations.
First, when the variables $y$ and $c$ influence different groups of features, meaning that $I^{\alpha} \neq I^{\beta} \neq I^{+} \neq I^{-}$. Then we have explored some cases in which the two variables affect the same group of features.

\begin{figure}
 \centering
  \includegraphics[scale=0.48]{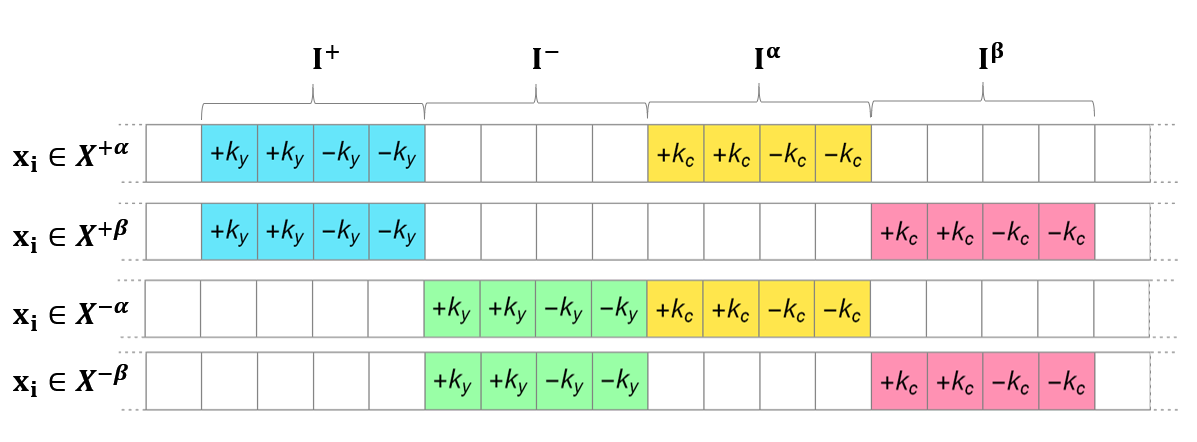}
  \caption{Schematic representation of the four subgroups of simulated data. Every feature vector is represented as a sequence of squares. All the squares (coloured or not) are affected by a random real noise in the range $[-10,10]$. The coloured ones represents the features that are influenced by $c$ and $y$, and this influence consists in the summation or subtraction of a constant from the noise.}
  \label{splitted_pos}
\end{figure}

\subsubsection{CI evaluation when the variables affect different features}\label{first_analysis_text}
In this analysis we consider the case in which $y$ and $c$ affect different features of the input data, as illustrated in Fig. \ref{splitted_pos}.\\
As shown in the figure, we assume that the constant added for $y=+1$ is the same for $y=-1$ and it is called $k_y$. Thus, the two classes differ only for the set of features depending on $y$ that are $I^{+}$ and $I^{-}$. The same happens also for the variable $c$, characterized by an additive constant $k_c$.\\
In this situation, in which $I^{\alpha} \neq I^{\beta} \neq I^{+} \neq I^{-}$, we want to study how our $CI$ responds to changes with respect to both $k_y$ and $k_c$.
Then, we want to assess whether the confounding effect of $c$ depends on which correlation between $y$ and $c$ we choose when building the training set, which means that we want to study the difference between $\Phi$ and $\Phi^*$.
Finally, we want to prove what we have previously said about the importance of calculating $\Phi$ (or $\Phi^*$) as the sum of $\Phi_{pro}$ and $\Phi_{cons}$ (or $\Phi^*_{pro}$ and $\Phi^*_{cons}$), showing how these quantities vary with respect to $k_c$ and $k_y$.\\
To perform these analyses we have calculated the values of $\Phi_{pro}$, $\Phi_{cons}$, $\Phi^*_{pro}$, $\Phi^*_{cons}$ and thus $CI$ on the simulated data just described, with $k_y$ and $k_c$ varying from 0 to 10 with steps of 0.5. We have chosen this range because 0 represents the situation of no-differences and 10 is the range of oscillation of our artificial noise.\\
Finally, for the sake of comparisons, we conducted the same simulated analysis using the permutation based confounding estimate, described in Section \ref{Elias}, to verify if our $CI$ addresses all the shortcomings of that method.

\subsubsection{CI evaluation when the variables can influence the same features}\label{second_simulated_analysis}
With this second analysis we want to explore the behaviour of our $CI$ in more complex situations, when $c$ and $y$ affect the same group of features and the dependence of the input data from them is expressed by four different constants $k_+$, $k_-$,$k_{\alpha}$ and $k_{\beta}$.\\
At the end of Section \ref{def_subsec}, we hypothesized that $\Phi$ and $\Phi^*$ can differ when, for example, $I^{\beta}=I^{+}$ and with this analysis we want to test our hypothesis and experimentally study other situations that can cause a difference between $\Phi$ and $\Phi^*$.\\
Exploring all the possible combinations of intersections between $I^{\alpha}$, $I^{\beta}$, $I^{+}$ and $I^{-}$ is clearly computationally infeasible, thus we analyze only the extreme situations represented in Table \ref{table_positions}, in which two or more of these groups of indexes are equal. In this table, a symbol denotes a specific group of features, while different symbols identify non-intersecting groups of features; thus, when two or more of $I^{\alpha}$, $I^{\beta}$, $I^{+}$ and $I^{-}$ have the same symbol, it means that those groups of features are influenced at the same time by two or more of the possible values of $(y,c)$.
We let the values of all the four constants vary in $\{-5,-1.5,0,1.5,5\}$, in order to explore the effects of small and large differences.

\begin{table}
\centering
 \begin{tabular}{||c c| c c||} 
 \hline
 $I^+$ & $I^-$ & $I^{\alpha}$ & $I^{\beta}$ \\ [0.5ex] 
 \hline\hline
 $\bigtriangleup$ & $\bigtriangleup$ & $\blacklozenge$ & $\blacklozenge$ \\
 \hline
 $\bigtriangleup$ & $\bigtriangleup$ & $\blacklozenge$ & $\square$ \\
 \hline
 $\bigtriangleup$ & $\blacklozenge$ & $\square$ & $\square$ \\
 \hline
 $\bigtriangleup$ & $\blacklozenge$ & $\square$ & $\bigstar$ \\
 
 \hline\hline
 $\blacklozenge$ & $\blacklozenge$ & $\blacklozenge$ & $\blacklozenge$ \\
 \hline
 $\blacklozenge$ & $\blacklozenge$ & $\blacklozenge$ & $\square$ \\
 \hline
 $\blacklozenge$ & $\blacklozenge$ & $\square$ & $\blacklozenge$ \\
 \hline
 $\blacklozenge$ & $\square$ & $\blacklozenge$ & $\blacklozenge$ \\
 \hline
 $\square$ & $\blacklozenge$ & $\blacklozenge$ & $\blacklozenge$ \\
 \hline
 $\blacklozenge$ & $\bigtriangleup$ & $\blacklozenge$ & $\bigtriangleup$ \\
 \hline
  $\blacklozenge$ & $\bigtriangleup$ & $\bigtriangleup$ & $\blacklozenge$ \\
 \hline\hline
 $\blacklozenge$ & $\square$ & $\blacklozenge$ & $\bigtriangleup$ \\
 \hline
 $\square$ & $\blacklozenge$ & $\blacklozenge$ & $\bigtriangleup$ \\
 \hline
 $\square$ & $\blacklozenge$ & $\bigtriangleup$ & $\blacklozenge$ \\
 \hline
 $\blacklozenge$ & $\square$ & $\bigtriangleup$ & $\blacklozenge$ \\
 \hline
\end{tabular}
 \caption{Table representing which groups of positions $I^+$, $I^-$, $I^{\alpha}$ and $I^{\beta}$ are equal in the analysis described in \ref{second_simulated_analysis}. Same symbols corresponds to same group of positions. For example, the first line represent the situation in which the dependency of the simulated data from $y=+1$ and $y=-1$ affects the same group of features (thus $I^+=I^-$), while the dependency from $c=\alpha$ and $c=\beta$ are expressed on the same group of features (thus $I^\alpha=I^\beta$), different from the previous one.}
 \label{table_positions}
\end{table}

\subsection{Neuroimaging Data}\label{neuroimaging_methods}
In this section, we describe an example of a possible application of our $CI$ on a real problem.
The problem taken as example is the classification of subjects affected by Autism Spectrum Disorders (ASD) versus Healthy Controls (HC), through the analysis of their neuroimaging data with ML algorithms.
Neuroimaging data, as every biological data, can possibly depend on a great number of phenotypical characteristics of the subjects, but the relationships that correlate the data to them are unknown. Thus, it is difficult both to apply a proper normalization and to decide for which variables it is essential to match the training subjects. \\
No standards exist in literature. Some studies take into account some characteristics that others are neglecting. In our analysis we will focus on the main phenotypical characteristics cited in literature as possibly confounding and we will show that our $CI$ gives us an idea of the importance of the problem, making the design of a study easier and more objective.\\
For this analysis we consider all the structural Magnetic Resonance Images (sMRI) available in the two collections ABIDE I \cite{di2014autism} and ABIDE II \cite{di2017enhancing}, the two biggest public databases for the study of ASD, containing brain magnetic resonance images of 2226 subjects.
These images have been processed with Freesurfer version 6.0 \cite{fischl2012freesurfer}, the most commonly used software for the segmentation of the brain. This processing extracts quantitative morphological features related to the cortical and subcortical structures and to the whole brain: these last ones are generally called global features. 
We selected 296 brain morphometric features, divided into:
\begin{itemize}
    \item volumes of 38 sub-cortical structures (the cortical structures are defined according to the Aseg Atlas\cite{fischl2002whole});
    \item 10 whole brain features;
    \item volume, surface area, mean thickness and mean curvature of 31 cortical bilateral structures, segmented according to the Desikan-Killiany-Tourville Atlas\cite{klein2012101}.
\end{itemize}
The analysis consists in testing the confounding effects of various variables, computing all the AUCs necessary for the calculation of our $CI$ with respect to the task of distinguishing between HCs and ASDs. 
As already explained in Section \ref{def_CI}, in order to obtain a good estimation of the $CI$ the bias added in the training set must be related exclusively to the variable under examination, while the others should be controlled. This means that, in order to obtain reliable results, the subjects of the two categories must be matched for all the variables that may be confounding and that are not under study. 
Furthermore, the matching operation must be performed on a case-by-case basis, i.e., for each ASD subject, another HC matched for all the possible confounding parameters, must be included in the training set.
This matching operation can be difficult to perform and unavoidably reduces the number of subjects that can be used for the training. However, in order to be able to compare the $CI$ calculated for the different possible confounding variables, it is important that all the training sets contain the same number of subjects and that the calculation of the various $CI$s are done exploring a sufficient number of biases.\\
The possibly confounding variables that we want to study are gender, age, handedness and Full Intelligence Quotient (FIQ). In fact, many authors supposed that the differences due to the mental abilities between ASDs and HCs can be a confounding factor that has to be avoided, because the meaning of finding a classifier able to distinguish between them is to help the physician to correctly diagnose which subject are ASDs and which ones are affected by other forms of mental retardation or neurodevelopmental disorders. \\
Besides these characteristics of the subjects that are typically mentioned as possibly confounding factors, we want also to analyze the $CI$ of the data acquisition site. In fact, ABIDE, as most of the neuroimaging datasets, is a multicentric database and its sMRIs have been acquired in 40 different sites, each one using its own machines and acquisition protocols.\\
In most neuroimaging studies, data are analyzed without taking into account their different acquisition modalities for two reasons. First, because usually the database collects data acquired with the same macroscopical sMRI settings, which are considered to produce equivalent images. Second, because data used in the classification task are not the raw image data that may depend on the acquisition settings, but features obtained with segmentation tools that are supposed to extract more abstract quantities, as stated by the authors of Freesufer \cite{fischl2004sequence}.\\
However, given the scarce reproducibility of the results obtained in neuroimaging literature, in the last years, an ever growing awareness has spread that the use of multicentric datasets may bias the analysis \cite{auzias2016influence}.\\
Summarizing, in this application we want to study the confounding effect of gender, age, handedness, FIQ and site in a classification problem between ASDs and HCs using neuroimaging data.
Age and FIQ are continuous variables, thus, in order to compute the $CI$ in these cases it is necessary to discretize their values. 
This has been done selecting the length $l$ of the range of values that represent a single discrete unit and confronting the value of $CI$ considering the confounding effect of two units separated by a distance $d$.
In particular, for the evaluation of the confounding effect of age, we chose to discretize the range at steps of $l=3$ years, starting from a group of 14 year old subjects. For the evaluation of the FIQ variable we instead considered $l=15$ points, starting from a FIQ score of 76.

\section{Results and discussion}
\subsection{Simulated Data}
\subsubsection{CI evaluation when the variables affect different features}\label{sim_diff_feat}
\begin{figure}
 \centering
  \includegraphics[scale=0.55]{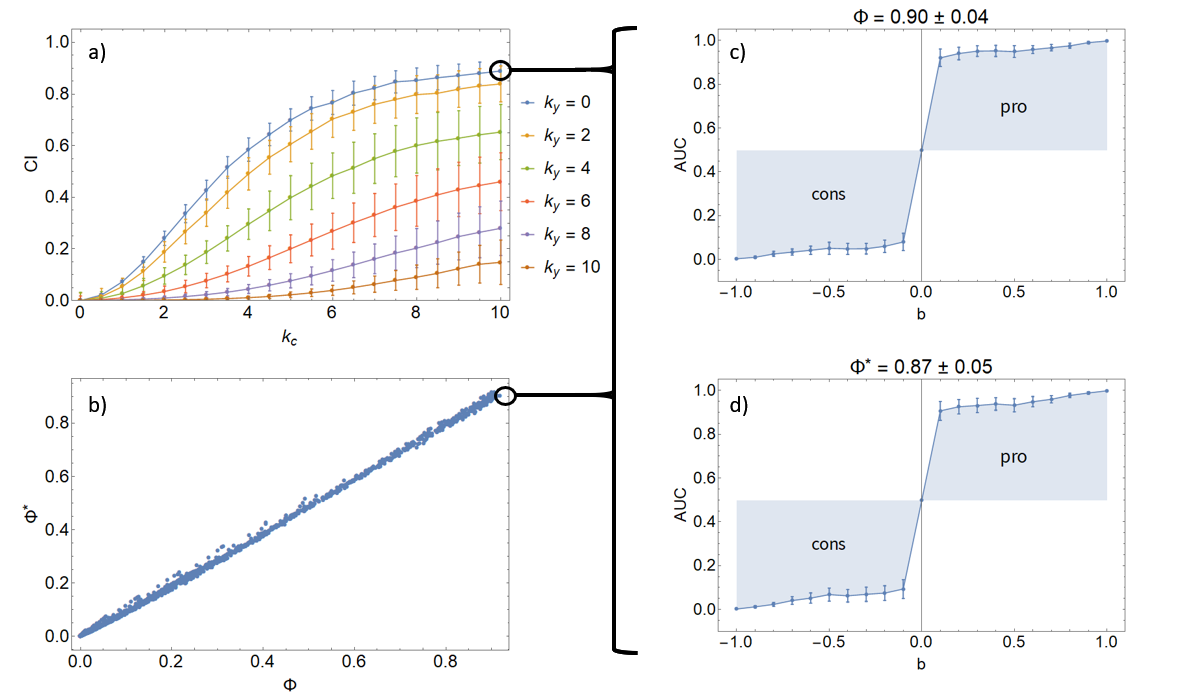}
  \caption{Results obtained from the analysis described in Section \ref{first_analysis_text}. Fig. ($a$) shows how the $CI$ calculated depends on $k_c$ and $k_y$. Fig. ($b$) shows along the two axes the values of the $\Phi$ and $\Phi^*$. All the points calculated with the simulated data of the first analysis lays in the line $\Phi=\Phi^*$. Fig. ($c$) and ($d$) show respectively the $\Phi$ and $\Phi^*$ computed for the definition of the $CI$ when $k_y=0$ and $k_c=0$. These plots are visual examples of how to calculate $\Phi$ and $\Phi^*$. They shows along the $x$ axes the values of the bias $b$ explored (negative values are the ones calculated with an unfavorable bias) and along the $y$ axes all the corresponding $AUCs$ computed. The light blue areas are the contributions that define $\Phi$ and $\Phi^*$. }
  \label{first_analysis}
\end{figure}

The results of the analysis described in Section \ref{first_analysis_text} are reported in Fig. \ref{first_analysis}a, in which the $CI$ values are plotted as a function of $k_c$ and $k_y$.

As the plot shows, our $CI$ depends both on $k_c$ and $k_y$.
Furthermore, as we would expect, the confounding effect of $c$ is weaker for easier tasks (i.e. the ones with higher $k_y$) and stronger for harder tasks.\\
The Fig. \ref{first_analysis}c and \ref{first_analysis}d are an example of how every point in Fig. \ref{first_analysis}a is calculated; in fact $CI$ is the maximum value between $\Phi$ and $\Phi^*$, the two quantities shown in the plots.
These quantities are calculated considering the two possible correlations between $y$ and $c$. They are obtained as the sum between two areas, one labelled $Pro$, that shows how much the $AUCs$ of a classifier increase if trained and validated on a favorably biased dataset, and the other called $Cons$ that on the contrary shows how they decrease when the bias is unfavorable.
For visualization purposes we attributed to the bias $b$ a negative value when computing the $AUCs$ of the $Cons$ part.\\
In these images $\Phi$ and $\Phi^*$ can be considered equal within the estimated error, and we have found that this is true also for all the $CI$s calculated in this simulation, in which $c$ and $y$ affect different features. This can be intuitively visualized from the Fig. \ref{first_analysis}b, in which the values of $\Phi$ and $\Phi^*$ are reported on the two axes. As this image shows, all the points lay on the line $\Phi^*=\Phi$, which means that the two values are consistent.\\

\begin{figure}
 \centering
  \includegraphics[scale=0.55]{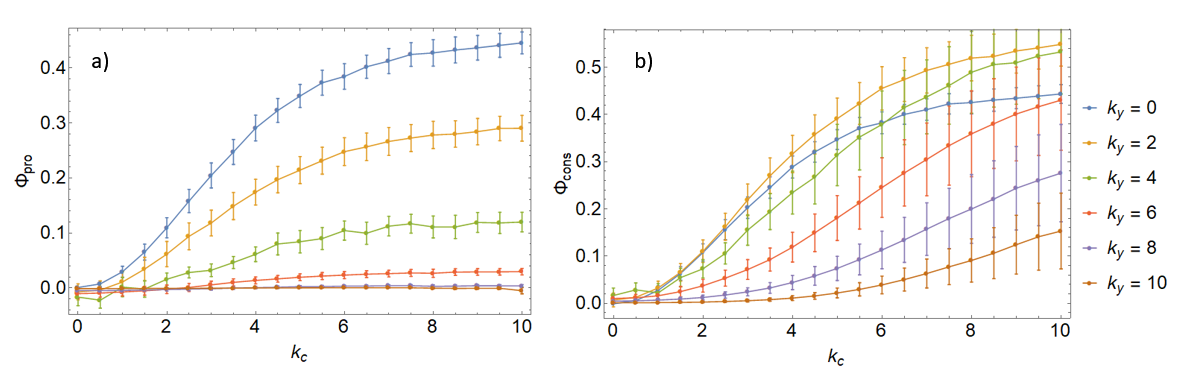}
  \caption{The images $a$ and $b$ show respectively the $\Phi_{pro}$ and $\Phi_{cons}$ obtained using simulated data with different values of $k_c$ and $k_y$. }
  \label{first_analysis_Cpro_Ccons}
\end{figure}

Note that in Fig. \ref{first_analysis}c and \ref{first_analysis}d, the $Cons$ and $Pro$ contributions seem equivalent, but these plots have been obtained when the $AUC$ in an unbiased situation is 0.5, i.e., $k_y=0$.  When considering all the $k_y$ and $k_c$ combinations explored in this analysis we see that the two contributions are different (see Fig.  \ref{first_analysis_Cpro_Ccons}).
In fact, when the tasks are easy, the unbiased $AUCs$ are already very high and thus a favorable bias cannot significantly improve them, resulting in too small $Pro$ contributions even for high values of $k_c$. In these cases in fact, the confounding effect is not manifested with an improvement of the $AUC$, but with a change in the classification pattern extracted during the training.
Similarly, when the tasks are hard, even a small unfavorable bias can significantly reduce the $AUC$, bringing it next to 0 and making the $Cons$ contribution solely dependent on the unbiased $AUC$, which will be lower the harder the task is.
The problems just mentioned are both caused by the dependency of the $Pro$ and $Cons$ contributions on the unbiased $AUC$. Their sum does not suffer from this dependency (see Eq. (\ref{C})) and it is, thus, the best figure of merit to correctly assess the entity of the confounding effect for any task complexity.\\

\begin{figure}
 \centering
  \includegraphics[scale=0.55]{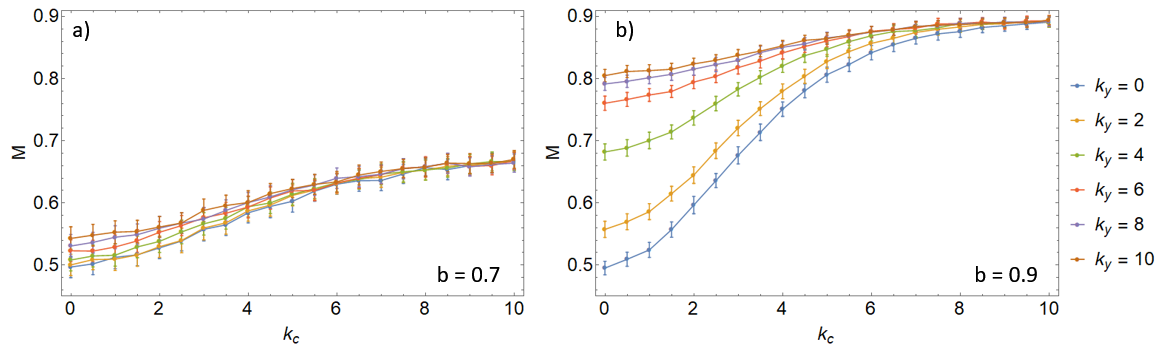}
  \caption{Results of the permutation based confounding estimate (described in Section \ref{Elias}) on the simulated data (described in Section \ref{first_analysis_text}). The analysis has been performed with bias $b=0.7$, Fig. a, and with $b=0.9$, Fig. b.}
  \label{Elias_comparison}
\end{figure}

Finally, we repeated the same simulation study for the permutation based confounding estimate described in Section \ref{Elias}, to show the main differences with respect to our $CI$.
First of all, while our $CI$ aims at determining the effect of a variable in a classification study, independently from the specific bias $b$ present in the dataset, this measure is strictly tied to the dataset composition. 
Fig. \ref{Elias_comparison} shows the distributions of the $M$ values computed for different $k_c$ and $k_y$ in the case of a bias $b=0.7$ (Fig. \ref{Elias_comparison}a) and in the case $b=0.9$ (Fig. \ref{Elias_comparison}b).
Both the figures show a differentiation with respect to the value of $k_c$, however it is less evident when the bias is small. Furthermore, as we already showed in Section \ref{Elias}, this method is not able to fully isolate the confounding effect. This causes a type I error that can be easily observed in Fig. \ref{Elias_comparison}b, because for $k_c = 0$ the measure $M$ increases with $k_y$. Our $CI$ instead is robust to this kind of error; as it can be noted in Fig. \ref{first_analysis}a, it  correctly estimates $c$ to be not confounding when $k_c = 0$ independently from the value of $k_y$. The inability of $M$ to disentangle the information learnt from the confounder from that learnt from the task, brings to another pitfall. Given the same bias $b$ and the same confounder strength $k_c$, the measure $M$ is higher when the task is easier (i.e. $k_y$ is greater), which is clearly incorrect. 
Our $CI$, instead, measures weaker confounding effects when the task is easier, as expected. 

\begin{figure}
 \centering
  \includegraphics[scale=0.55]{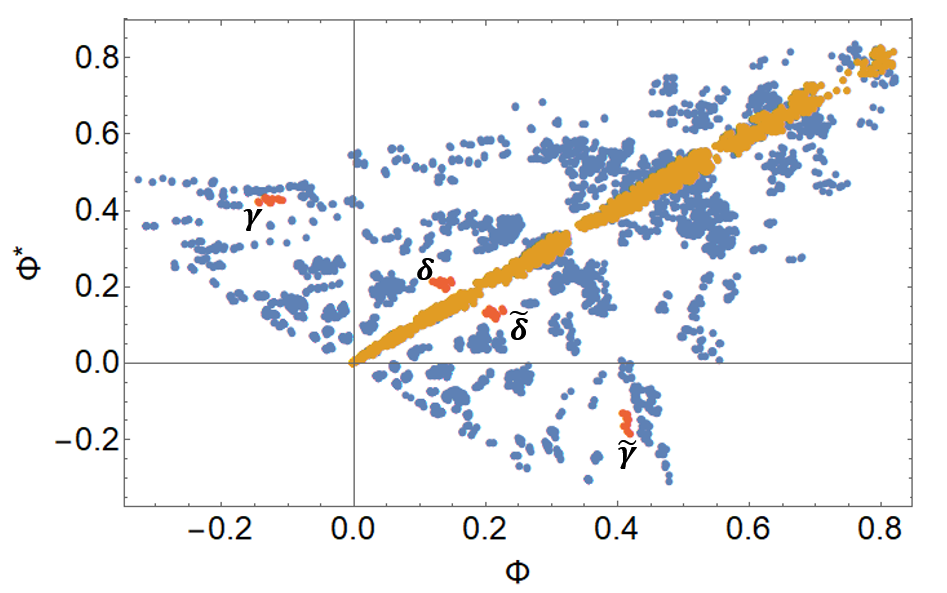}
  \caption{Plot of the $\Phi^*$ and $\Phi$ obtained with the analysis described in Section \ref{second_simulated_analysis}. The orange points have been obtained with simulated data in which the features affected by $y$ are different from the ones affected by $c$. The blue points seems organized in symmetrical clusters, of which some examples analyzed in the text are coloured in red.}
  \label{second_analysis}
\end{figure}

\subsubsection{CI evaluation when the variables can influence the same features}
\begin{figure}
 \centering
  \includegraphics[scale=0.48]{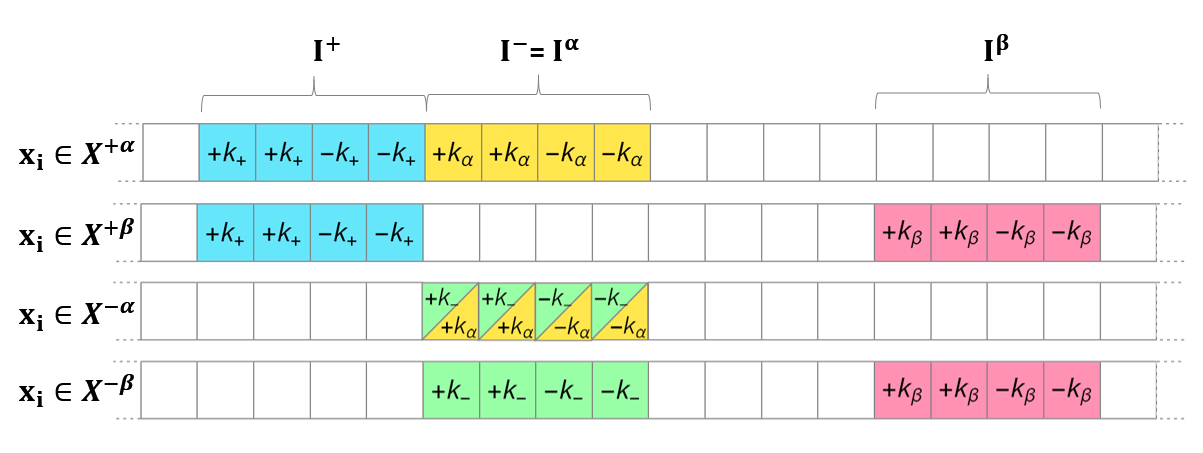}
  \caption{Figurative representations of the simulated data used for the calculation of the $\Phi$ and $\Phi^*$ values depicted in Fig. \ref{mixed_cluster} and \ref{positive_cluster}}
  \label{positions_cluster}
\end{figure}

\begin{figure}
 \centering
  \includegraphics[scale=0.75]{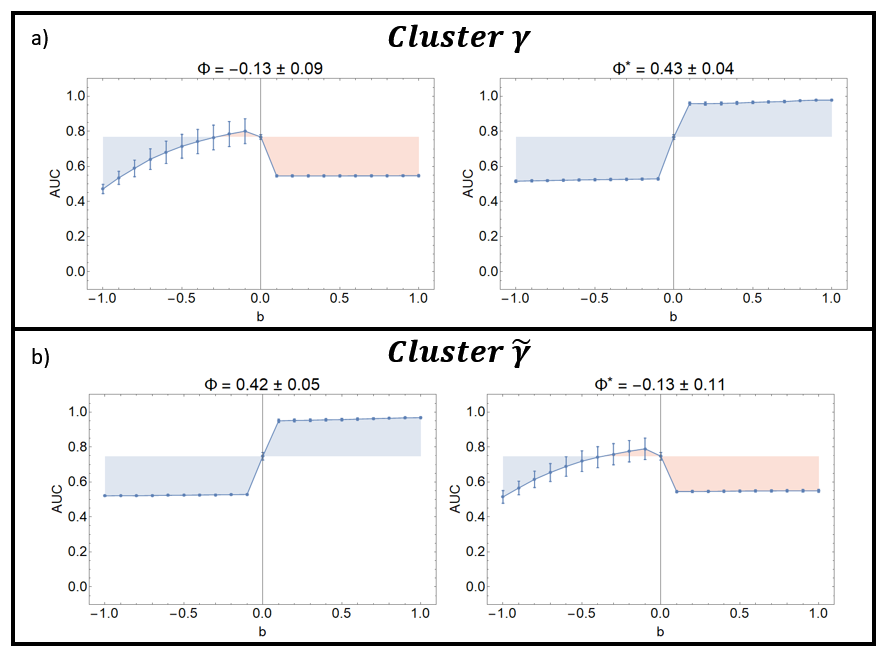}
  \caption{a) Plots of the AUCs used to compute the values of $\Phi$ and $\Phi^*$ for the cluster $\gamma$ in which $(k_+,k_-,k_{\alpha},k_{\beta})= (-1.5,5,-5,0)$ and $I^{-} = I^{\alpha}$. This means that the effects of $k_-$ and $k_\alpha$ (which influence the same features) nullify each other when applied to the same feature vector, thus making $\Phi$ (in which $y=-1$ and $c=\alpha$ are positively correlated) useless as an estimator of the confounding effect of $c$. This shows the necessity of calculating both $\Phi$ and $\Phi^*$ and to analyse the monotonicity of the AUCs used to calculate them (in fact, the curve in the plot of $\Phi$ is clearly not monotone).
  b) Same plot of Fig. a), but with $(k_+,k_-,k_{\alpha},k_{\beta})= (-1.5,-5,-5,0)$. This time the effects of $k-$ and $k_\alpha$ augment each other when applied to the same feature vector, thus resulting in an opposite situation with respect to Fig. a).}
  \label{mixed_cluster}
\end{figure}

\begin{figure}
 \centering
  \includegraphics[scale=0.75]{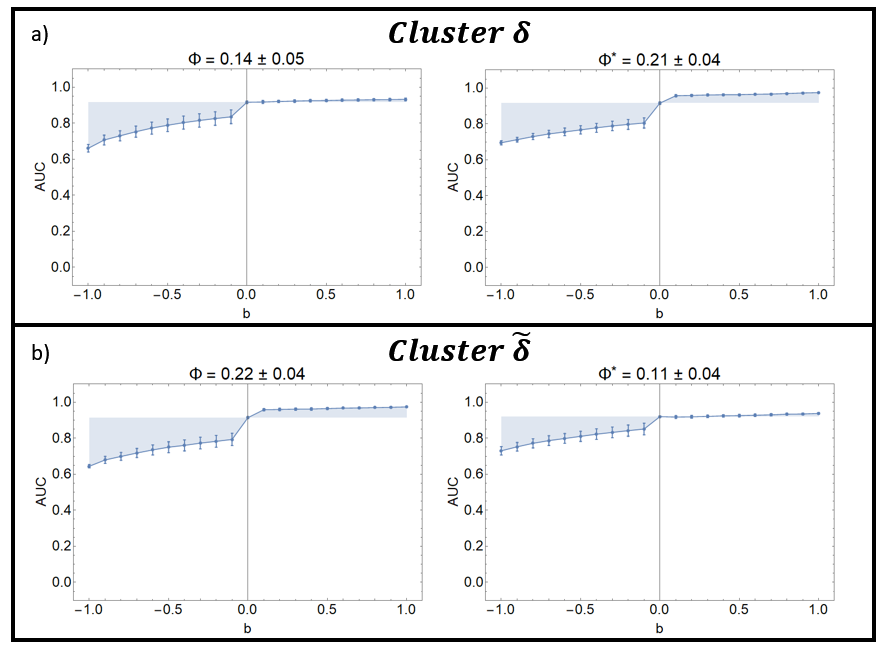}
  \caption{a) Similar to Fig. \ref{mixed_cluster}, but with a smaller $|k_\alpha|$: $(k_+,k_-,k_{\alpha},k_{\beta})= (5,5,1.5,5)$. This time $k_-$ and $k_\alpha$ do not cancel each other when applied together, but still their effect is diminished. This can be noted in the plots since $\Phi^*>\Phi$.
  b) Same plot of Fig. a), but with $(k_+,k_-,k_{\alpha},k_{\beta})= (5,5,-1.5,5)$. The effects of $k-$ and $k_\alpha$ augment each other when applied to the same feature vector, thus resulting in an opposite situation with respect to Fig. a).}
  \label{positive_cluster}
\end{figure}

Similarly to Fig. \ref{first_analysis}b, Fig. \ref{second_analysis} shows the results obtained with the analysis described in Section \ref{second_simulated_analysis}, showing along the two axes the values of $\Phi$ and $\Phi^*$. The orange points, that all lay in the diagonal of the plot, have been obtained with the first 4 configurations described in Table \ref{table_positions}. All of them are characterized by the absence of intersections between the group of features influenced by $y$ and the ones influenced by $c$: $(I^+ \cup I^-) \cap (I^{\alpha} \cup I^{\beta}) = \varnothing$.
As expected, the correlation between the values of $c$ and $y$ chosen for the calculation of the index is irrelevant, and thus $\Phi = \Phi^*$, if $c$ and $y$ affect different features.\\
The other blue points seem to be grouped in clusters that are symmetrical with respect to the orange line. In these clusters, we can identify two kinds of situations: the one in which both $\Phi$ and $\Phi^*$ are positive (first quadrant), and the one in which one of the two quantities is positive while the other one is negative (second and third quadrant).\\
To better understand when these situations occur, we have analyzed the two clusters (and their respective symmetrical ones) coloured in red.\\
The points belonging to the clusters in the second and third quadrants, $\gamma$ and $\tilde{\gamma}$, are all obtained with data simulated in a configuration in which one specific value of $c$ and one specific value of $y$ influence the same features, while the other values affect different and independent groups of features (see the last four lines in Table \ref{table_positions}). Elements of a specific cluster (and its symmetrical one) have been obtained with the same module of $k_+$, $k_-$, $k_{\alpha}$, $k_{\beta}$.\\
The only difference between two symmetrical clusters, e.g., $\gamma$ and $\tilde{\gamma}$, is the signs of the $k$ constants affecting the same features. In the case depicted in Fig. \ref{positions_cluster}, these signs are those of $k_-$ and $k_\alpha$.\\
To better understand why in these situations either $\Phi$ or $\Phi^*$ assumes a negative value, let us consider the plots in Fig. \ref{mixed_cluster}a and \ref{mixed_cluster}b, showing the components of two symmetrical points belonging respectively to the $\gamma$ and $\tilde{\gamma}$ clusters.
The input data used for the calculation of the quantities in Fig. \ref{mixed_cluster}a have $I^- = I^{\alpha}$ as illustrated in Fig. \ref{positions_cluster} and are characterized by the constants $(k_+,k_-,k_{\alpha},k_{\beta})= (-1.5,5,-5,0)$. Thus, when there is a positive correlation between the variable $y=-1$ and $c=\alpha$, their effects cancel each other in a significant portion of the training dataset. This results in a negative value of $\Phi$. Instead, when considering a positive correlation between $y=+1$ and $c=\alpha$, the confounding effect due to $c$ is even greater with respect to the situation in which $I^{\alpha} \neq I^{\beta} \neq I^{+} \neq I^{-}$ (with the same $k$ values). This happens because, if the features belonging to $I^{-} = I^{\alpha}$ are affected by $k_-=5$ when $x_i \in X^-$ and by $k_{\alpha}=-5$ when $x_i \in X^{\alpha}$ and $X^{\alpha}$ is positively correlated with $X^+$, it is like $C^*$ is measuring a hypothetically confounding factor with an intensity given by the difference of $k_-$ and $k_{\alpha}$, thus of $|5-(-5)|=10$.\\
For similar reasons, the symmetrical point has a negative $\Phi^*$ and a positive $\Phi$ as illustrated in Fig. \ref{mixed_cluster}b. In this case $(k_+,k_-,k_{\alpha},k_{\beta})= (-1.5,-5,-5,0)$, thus $\Phi$ measures the effects given by the sum of $k_-$ and $k_{\alpha}$, i.e. $|(-5)+(-5)|=10$. When computing $\Phi^*$, the training dataset is biased in a way that makes the greatest part of the $x_i \in X^+$ belonging also to $X^{\alpha}$, thus presenting the same effect on the same features that characterize the $x_i \in X^-$. This makes the data indistinguishable with respect to $y$ and brings the $AUCs$ of $\Phi^*_{pro}$ to 0.5.
These two examples show clearly what happens to all the points belonging to these clusters, explaining their symmetry and why there are cases in which $\Phi$ and $\Phi^*$ can differ.
However, it is interesting that even in these very degenerate cases, our index gives the same estimation of the confounding effect, correctly reflecting the fact that they are characterized by the same absolute values of the $k$ constants. This happens because $CI$, being computed as the maximum value between $\Phi$ and $\Phi^*$, always considers the worst possible correlation between $y$ and $c$.\\
Something similar happens also for the clusters of the first quadrant $\delta$ and $\tilde{\delta}$, that are composed of points obtained with the same configurations of $\gamma$ and $\tilde{\gamma}$, but with $k$ values that make the task easier and with $|k_\alpha|<|k_-|$. For example, let us consider the two points belonging respectively to $\delta$ and $\tilde{\delta}$ illustrated in Fig. \ref{positive_cluster}; they have been obtained from simulated data with constants $(k_+,k_-,k_{\alpha},k_{\beta}) = (5,5,\pm1.5,5)$. In this situation, the difference between $\Phi$ and $\Phi^*$ is less important than in the one represented in Fig. \ref{mixed_cluster} because the task is easier and because $|k_\alpha|<|k_-|$ implies that their effects do not cancel each other completely when they are correlated.\\
Concluding, our analysis on these clusters shows that the situations in which one of $\Phi$ or $\Phi^*$ is negative are not inherently different from the situations in which they are both positive but different.
We also believe that the cluster distribution is simply an effect of the coarse grained sampling of the $k$ values and of neglecting partial intersections between the groups of features affected by $y$ and $c$. Despite not being able to explore all the possible cases described above, it is interesting to note that in none of the performed analyses $\Phi$ and $\Phi^*$ were both negative.

\subsection{Neuroimaging Data}
\begin{figure}
 \centering
  \includegraphics[scale=0.35]{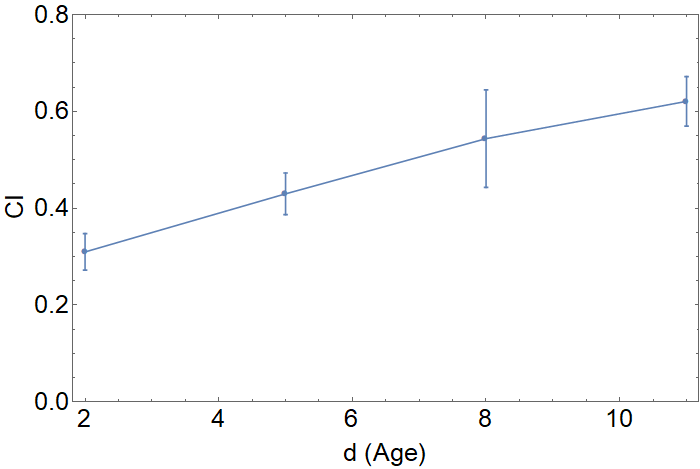}
  \caption{Plot of $CI$ as a function of the age difference $d$ between the two groups of subjects used in the analysis. The $CI$ shows that the higher the age difference is, the higher its confounding effect is.}
  \label{AGE}
\end{figure}

\begin{figure}
 \centering
  \includegraphics[scale=0.35]{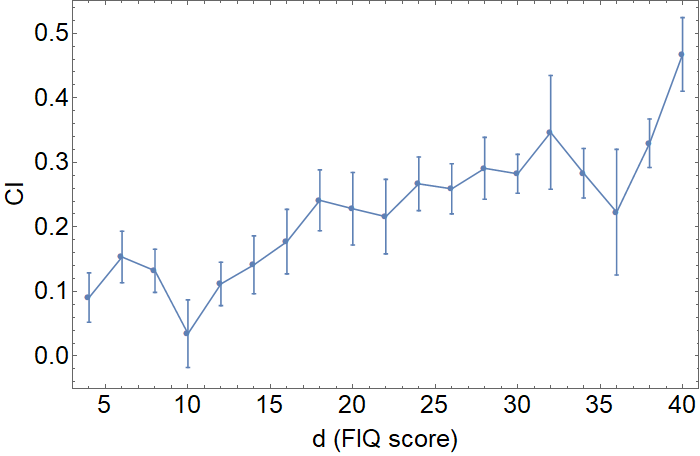}
  \caption{Plot of $CI$ as a function of the FIQ score difference $d$ between the two groups of subjects used in the analysis. The $CI$ shows that up to about 15 points, the difference in $CI$ can be classified as almost non-confounding, while it becomes an important factor for higher values of $d$.}
  \label{FIQ}
\end{figure}

The $CI$s calculated in this analysis are reported in Table \ref{table_CI} for the categorical variables and in Fig. \ref{AGE} and \ref{FIQ}, for the continuous ones, respectively age and FIQ.
The discretization strategy for the age and FIQ variables is described in Section \ref{neuroimaging_methods}. The first point of the plot in Fig. \ref{AGE}, in which $d=2$, shows the $CI$ calculated considering as $X^{\alpha}$ and $X^{\beta}$ two groups of subjects with an age in the ranges $14-17$ years and $19-21$ years respectively. Even considering that these two ranges are very close, the $CI$ is sensibly different from 0. As we would expect, the value of $CI$ increases with $d$, exceeding 0.6 for $d = 11$.
The $CI$s of the FIQ variable are smaller than the $CI$s of age and this is also reflected by the oscillations present in Fig. \ref{FIQ}. However, an increasing trend is clearly detectable, bringing the $CI$ of FIQ to be significant for high values of $d$. It is remarkable that the points that seem out of the trend are the ones affected by the greatest error.\\
Summarizing, the application here reported is just an example of how the $CI$ can be used to understand better the best conditions to design a ML study in the presence of confounding variables. Clearly, to calculate the CI it is necessary to reduce the number of subjects under examination, in order to match all the possible confounding factors. For example, the assessment of the confounding effect of the sex variable required simultaneously matching for FIQ, site and age. This matching operation generated a dataset of about 150 subjects. It is important to note that the sex variable is particularly unfortunate since the number of ASD females in the whole ABIDE dataset is only 142.
However we believe that the CI calculation can help the data analyst to optimize the number of subjects to include in the final analysis, providing a way to objectively evaluate which variables are more confounding (and thus must be matched in training). This analysis, for example, shows that the handedness category is not a confounding variable for the task under examination. Even if these results are related to the features and the specific classifier chosen, we believe that in many studies, recruitment choices such as reducing the dataset to only right-handed subjects, have unduly limited the subjects cohort available for training without a valid justification.\\
On the other hand, many studies have completely neglected the dependency of the data from the acquisition modalities and from the FIQ. This last variable can be very important, especially if the disease is related to mental disability, while the HCs follow a normal distribution.
This study also shows that the $CI$ of the acquisition site is higher than the one related to gender and comparable to the one of an age difference of about 11 years, and thus should not be underestimated in a multicentric study. \\
It is interesting that in all these examples we have never obtained a $CI$ significantly higher than 0.6, even if, for example, simply training the logistic regression classifier to distinguish between matched subjects acquired from two different sites we obtain a mean AUC of $0.99 \pm 0.03$. This is due to the fact that the $CI$ reaches values around 1 only when the confounding effect is so powerful to completely mislead the classifier even with very small biases in the training set; this rarely happens with real data.

\begin{table}
\centering
 \begin{tabular}{|c| c c|} 
 \hline
 Variable & $CI$ & Error \\ [0.5ex] 
 \hline\hline
 Handedness & 0.01 & 0.06 \\
 \hline
 Sex & 0.43 & 0.03 \\
 \hline
 Site & 0.54 & 0.02 \\
 \hline
\end{tabular}
 \caption{CI values and their estimated error for the categorical variables under examination in our application on neuroimaging data.}
 \label{table_CI}
\end{table}

\section{Conclusions}
In this paper we have presented an index for assessing the confounding effect of a categorical variable in a binary classification study.\\
The study made on simulated data shows the goodness and sensitivity of our CI, the value of which depends on the intensity with which the confounder and the label influence the features under exam.
Furthermore, it has been found that $\Phi$ and $\Phi^*$ differ only when $c$ and $y$ influence the same features. This phenomenon could give precious insights on how the confounder and the class label affect the data.\\
The analysis conducted on neuroimaging data showed very informative results, proving that the CI can also be used on continuous variables by discretizing their values.
The analyses on real and simulated data were aimed at proving the goodness of the CI as a measure to rank the effect of multiple confounders. Nonetheless, this figure of merit can be used also to assess the effectiveness of a normalization procedure or of a learner algorithm specifically designed to be robust against confoundings.\\
In this paper we have also discussed the limits and the differences with the only other work that, with similar aims, presented a method for measuring the confounding effect in a classification study. In fact, we have shown that this method commits an error of I type and provides a measure that heavily depends on the bias in the dataset. Our index, instead, is robust against that error type and is independent from the dataset composition, since it captures how strongly the confounder affects the data with respect to the complexity of the task.\\
Concluding, the proposed CI represents a novel and robust instrument to measure confounding effects and evaluate the effectiveness of possible countermeasures. It can be used for various known confounding problems, especially in the biomedical sector, such as:
\begin{itemize}
    \item demographic characteristics in imaging \cite{rao2017predictive, brown2012adhd};
    \item subject identification in longitudinal digital health \cite{neto2019detecting} and in augmented data segmentation problems \cite{wang2019removing};
    \item acquisition modalities, for any study based on the analysis of high dimensional data (such as MRI \cite{yamashita2019harmonization}, radiography \cite{zech2018variable}, gene expression \cite{parker2014removing}, etc.);
    \item head motion in MRI based disease recognition studies, especially for disorders affecting movement \cite{yendiki2014spurious};
    \item instructions to participants in a resting-state MRI study (i.e. open/close eyes \cite{yan2009spontaneous}).
\end{itemize}
In fact, medical data depend on many different variables and correcting an analysis from multiple confounding effects, without losing the signal of interest is not straightforward.

\bibliographystyle{model1-num-names}
\bibliography{sample.bib}

\end{document}